\def\year{2022}\relax
\documentclass[letterpaper]{article} 
\usepackage{aaai22}  
\usepackage{times}  
\usepackage{helvet}  
\usepackage{courier}  
\usepackage[hyphens]{url}  
\usepackage{graphicx} 
\usepackage{natbib}  
\usepackage{caption} 
\DeclareCaptionStyle{ruled}{labelfont=normalfont,labelsep=colon,strut=off} 
\usepackage{algorithm}
\usepackage{algorithmic}

\usepackage{newfloat}
\usepackage{listings}
\usepackage{booktabs}
\usepackage{multirow}
\floatstyle{ruled}
\newfloat{listing}{tb}{lst}{}
\floatname{listing}{Listing}
\title{Learning Modular Structures That Generalize Out-of-Distribution\\(Student Abstract)}
\author{
    Arjun Ashok,
    Chaitanya Devaguptapu,
    Vineeth N Balasubramanian
}
\affiliations{
    Indian Institute of Technology, Hyderabad\\
    IITH Main Road, Near NH-65, Sangareddy, Kandi, Telangana, India - 502285\\
    arjun.ashok@cse.iith.ac.in,
    cs19mtech11025@iith.ac.in, 
    vineethnb@cse.iith.ac.in

}

\begin{document}

\maketitle

\begin{abstract}
Out-of-distribution (O.O.D.) generalization remains to be a key challenge for real-world machine learning systems. We describe a method for O.O.D. generalization that, through training, encourages models to only preserve features in the network that are well reused across multiple training domains. Our method combines two complementary neuron-level regularizers with a probabilistic differentiable binary mask over the network, to extract a modular sub-network that achieves better O.O.D. performance than the original network. Preliminary evaluation on two benchmark datasets corroborates the promise of our method.
\end{abstract}

\section{Introduction}

\noindent Recent work has uncovered that neural networks that are learned on observational data are often prone to spurious correlations, and rely on shortcuts learned from the training data for solving the task instead of modelling the underlying mechanism \cite{Geirhos_2020}. This leads to them failing to transfer to more challenging testing conditions, such as real-world scenarios. Recent works show that modularity is a useful inductive bias that can lead to better systematic generalization \cite{goyal2020recurrent, csordas2021are}. We seek to understand whether networks can be structurally enforced to prefer modular solutions. In this context, \cite{zhang2021subnetwork} show that a fully-trained network contains sub-networks that are less susceptible to spurious correlations, and introduce a method to extract the structure from a trained network. We study whether we can obtain such solutions through training itself, by regularizing the network to avoid fitting spurious correlations in the data. We introduce objectives that explicitly incorporate the structure of the network and induce modular structures to be formed at every layer of the network. Our method enforces the network to be a \textit{compositional hierarchy of expert modules}, promoting the emergence of features that are specialized and reused across multiple training domains in the network. We show that our method boosts the O.O.D. performance of networks across two benchmark datasets.

\section{Method}

\noindent We first motivate our method. A deep neural network contains several layers of neurons, each serving as a feature for every neuron in the next. Every fundamental sub-function (e.g. a single convolutional filter) in the network is associated with separate neurons that arise out of transformations (e.g. dot product) of the function with the input. Our aim is to discover features that are well used (activated) across multiple training domains, as well as detecting and preventing redundant features from being present in the network. Hence, \textit{specialization} and \textit{reuse} are two key principles that underlie our method that we describe below.

\bigskip
First, our objective for specialization regularizes such that 
\textit{\textbf{every feature in the network should be a different composition of the available sub-features}}. That is, every feature should fit as few features as necessary, and should differ as much as possible in the features fit, minimizing redundancy. However, directly encouraging this on the weights would unnecessarily constrain the power of the network.

\bigskip
Hence, we use a differentiable probabilistic binary mask $\pi_i$ over the network weights, relaxed by the Gumbel-Sigmoid estimator \cite{jang2017categorical}. Each value $\pi_i \in [0, 1]$ represents the probability of sampling weight $w_i$. During every training iteration, once the mask is sampled, it is binarized as $m_i = \{$sigmoid$(\pi_i) > 0.5\}$ $\in \{0,1\}$. Once trained, we obtain deterministic masks by binarizing the final values, hence extracting a subnetwork described by the mask.

\bigskip
As part of our \textit{specialization} objective, we impose the following regularization of the continuous masks of the weights:

\begin{equation}
\label{eqn:specialize}
S_1(\pi) = \sum_{l=1}^{L} \sum_{p=1}^{N_l} \left(\sum_{i=1}^{M_p} \pi_i\right)^2
\end{equation}

\bigskip
\noindent where $L$ denotes the number of layers, $N_l$ - the number of features in the layer, $M_p$ - the number of outgoing weights from feature $p$.

Note that we minimize the (square of) sum of sampling probabilities of weights outgoing from \textbf{each feature} in the current layer, allowing it to be \textit{fit sparsely} by only a few required features from above. Consequently, this encourages features in the next layer to fit a minimally overlapping set of features from the current layer, leading to each of the former specializing in their underlying function.

\bigskip
Although this objective would encourage specialization, every feature in the current layer may not be necessary, as extra features may correspond to unnecessary functions. The network must automatically be able to decide how many features to keep. However, constructing an objective that can be used to restrict the number of features in a layer is non-trivial, since in the worst case, every feature may be necessary depending on the task and model capacity at hand.

\bigskip
Here, we hypothesise that \textit{\textbf{the necessary features are those that are reused across domains by multiple specialist functions above}}. Consequently, we regularize to preserve only those features that have a large number of outgoing weights sampled with high-probability, discarding features that are not well reused. We enforce this through the following objective:

\begin{equation}
\label{eqn:prevent}
S_2(\pi) = \sum_{l=1}^{L} \sum_{p=1}^{N_l} \sqrt[]{\sum_{i=1}^{M_p} \pi_i^2}
\end{equation}

\bigskip
This term is inspired from that of group lasso regularization \cite{KimTree}; applying this term can effectively zero out the masks of \textit{all} the outgoing weights of some features. Unlike that of group lasso that regularizes the weights and can have overlapping groups, we apply it on the masks and do not have any overlapping groups.

\bigskip
Our final objective is, therefore,
$$L = \ell(\theta) + R(\theta) + \alpha*S_1(\pi) + \beta*S_2(\pi)$$ where $\ell$ is the loss function used for the task, $R$ being a general-purpose regularizer (eg. $L_2$), and $\alpha$ \& $\beta$, the weights of each of our regularization terms.

\bigskip
Training with the regularized differentiable mask on data consisting of multiple training domains conditionally activates only those weights shared across multiple domains. Consequently, the sub-network contains features that are invariant to the domain, and hence aids O.O.D. generalization.


\section{Preliminary Results}
\noindent We present preliminary results of our method on two benchmark O.O.D. generalization datasets - Colored MNIST (C-MNIST) and Rotated MNIST (R-MNIST). Each dataset is artifically biased in such a way that in the training dataset, a certain degree of correlation is induced between spurious variables and the class label. In the test dataset, the correlation is reversed. The goal of O.O.D. generalization is to encourage the model to fit the invariant features, ignoring other correlated variables, training and validating only on in-distribution data. 

\begin{table}[t!]
\centering
\begin{tabular}{c|l|cc}
\toprule
\multicolumn{1}{l|}{\textbf{Model}} & \textbf{Method} & \textbf{C-MNIST} & \textbf{R-MNIST} \\ \toprule
\multirow{4}{*}{CNN}                       & ERM             & 35.23                & 96.5                \\
                                           & ERM + modReg    & \textbf{38.20}                & \textbf{96.7}                \\ \cline{2-4}
                                           & IRM             & 67.69                & 97.3                \\
                                           & IRM + modReg    & \textbf{71.88}                & \textbf{98.1}                \\ \midrule
\multirow{4}{*}{MLP}                       & ERM             & 34.27                & 94.45                \\
                                           & ERM + modReg    & \textbf{36.91}                & \textbf{95.43}                \\ \cline{2-4}
                                           & IRM             & 72.58                & 97.4            \\
                                           & IRM + modReg    & \textbf{75.59}                & \textbf{97.9}                \\ \bottomrule
\end{tabular}
\caption{Results of the proposed method on multiple architectures, across datasets.}
\label{table:results}
\end{table}

Our method is versatile, and can be used on top of any algorithm. Here, we apply our method on top of empirical risk minimization(ERM), the standard approach to machine learning problems, and invariant risk minimization (IRM) \cite{arjovsky2020invariant}, a method that estimates invariant, causal predictors from multiple training environments.

\bigskip
Preliminary results shown in table \ref{table:results} verify the effectiveness of our method. Our method gives consistent gains across the two datasets and architectures considered. In particular, our gains give considerable boosts in the heavily biased C-MNIST dataset, and also improves performance in the R-MNIST dataset in which existing methods have reached their potential. 

\section{Future Work}

\noindent We plan to scale up our method and test its effectiveness on larger datasets. Further, we also plan to take our method forward and evaluate on larger architectures such as ResNets, and on top of other existing O.O.D. generalization methods. 


\bibliography{aaai22.bib}

\end{document}